# Grading of Mammalian Cumulus Oocyte Complexes using Machine Learning for *in vitro* Embryo Culture

Viswanath PS[†], *Graduate Student Member, IEEE,* Tobias Weiser[†], Phalgun Chintala[†], Subhamoy Mandal*, *Graduate Student Member, IEEE,* and Rahul Dutta*

*Abstract*— **Visual observation of Cumulus Oocyte Complexes provides only limited information about its functional competence, whereas the molecular evaluations methods are cumbersome or costly. Image analysis of mammalian oocytes can provide attractive alternative to address this challenge. However, it is complex, given the huge number of oocytes under inspection and the subjective nature of the features inspected for identification. Supervised machine learning methods like random forest with annotations from expert biologists can make the analysis task standardized and reduces inter-subject variability. We present a semi-automatic framework for predicting the class an oocyte belongs to, based on multi-object parametric segmentation on the acquired microscopic image followed by a feature based classification using random forests.**

## I. INTRODUCTION

*In vitro* production of embryo in animals is heavily dependent on a supply of slaughterhouse ovaries. Typically a mammalian ovary (including porcine species) contains about 4 to 5 million primordial follicles at 18 - 22 weeks post-conception. The supply of follicles decreases slightly before birth and keeps on depleting progressively throughout the lifespan of the animal. During the follicular phase, under hormonal influence a small cohort of follicles (approximately 10–14) begins to develop. One member of this group is physiologically selected to ovulate in uniparous mammals, exhibiting greatly increased hormonal activity and increasing its growth the others, lacking hormonal support, undergo atresia [1]. However, for *in vitro* culture, oocytes are isolated from a cohort of dominant and subordinate follicles, most commonly by aspiration. Follicles of size roughly 3-6 mm are chosen for aspiration simply based on the visual observation. The developmental competence of the *in vitro* isolated oocytes varies according to the developmental stage of the ovarian follicle. As a result it leads to a batch-to-batch variability during *in vitro* embryo production. In all mammalian species including human, cumulus oocyte complexes (COCs) are usually graded by visual assessment of morphological features such as the thickness and compactness of the cumulus investment, ooplasm homogeneity [2] and the size of follicles [3] or oocytes [4]. Based on light microscopy, mammalian COCs surrounded by several layers of cumulus cells and with evenly granulated ooplasm have higher developmental competence in vitro than oocytes with irregularly granulated ooplasm and fewer cumulus layers[5].

In spite of the different selection criteria that have been reported, visual observation provides only limited information about its functional competence. Alternative methods of analysing cumulus cells are available, including evaluation of telomere length, cumulus cell apoptosis and gene expression profiling using microarray analysis, but are too time consuming and expensive for routine laboratory use. The cumulus layer is undoubtedly vital for oocyte development. For example, it plays an important role in the distribution of cortical granules, and thus the ability to undergo sperm penetration. Propidium iodide staining has also been used as an invasive method of determining cumulus cell integrity, but this requires fluorescent microscopy, which may not be available for some researchers and is potentially mutagenic. A simpler non-invasive assay for oocyte quality control, based on supravital dye lissamine dye, was recently proposed [6]. Since all of these oocyte grading methods, based on morphological evaluations are subjective and categorisation standards vary among investigators standard assays for quality control are needed for better standardisation of *in vitro* embryo culture. Therefore, an artificial intelligence approach is the need of the hour. In literature, there are several reference of utilizing image analytics for evaluating embryo development tracking and selection, however, work dedicated solely towards oocyte selection is limited. The articles illustrating the application on machine learning towards improvement of performance of in-vitro fertilization (IVF) focusses on exaction of morphology from images of embryos and grading them before transfer. Manna et al [7] has proposed a textural image descriptor (Local binary patterns (LBP)) followed by a decision support system to identify viable oocytes and embryos for IVF. However, biologically relevant features such as relation between the various cumulus layers have not yet been explored.

In the current article, we propose a newer approach wherein we use parametric segmentation for delineate the oocyte complexes, thereafter use a bag of features to train an ensemble of decision trees. The forest of decision trees thus created can successfully grade oocyte into various grades as mandated by embryologists. This approach essentially allows the automation of the process (Figure 1) for selection of

[†] Authors contributed equally towards the work.

Vishwanath PS, Phalgun Chintala are with the Faculty of Informatics and Mathematics, Technische Universität München, Boltzmannstraße 3, 85748 Garching b. München, Germany.

*S Mandal is with the Faculty of Electrical Engineering & Information Technology, Technische Universität München, and Institute of Biological and Medical Imaging, Helmholtz Zentrum München, Ingolstadter Landstr 1, 85764 Oberschleissheim, Germany. (e-mail: s.mandal@tum.de).

Tobias Weisser and *Rahul Dutta, are with Chair of Livestock Biotechnology, Technische Universität München, Freising, Germany. (e-mail: doctordut@gmail.com; Ph. +49 (0) 8161-712032).

competent oocytes which positively impacts the performance of IVF programs. The emphasis is on image features which are relevant to the current problem and to reduce the time complexity with reasonable accuracy.

## II. METHODS AND ALGORITHM

### A. Imaging set up for oocyte collection and Grading

Slaughterhouse ovaries of prepubertal gilts were brought to the laboratory within one hour of collection in a temperature-controlled box maintained at 39º C. Ovaries were thoroughly washed with pre-warmed (39º C) Dulbecco's phosphate-buffered saline (DPBS) solution containing 0.1% polyvinyl alcohol and penicillin, streptomycin solution. For collection of oocytes, oocytes collection medium consisting of TCM-199, 25 mM HEPES, L-glutamine, BSA, gentamicin and 10% FBS was used. Fresh medium (100 ml) was prepared, filtered through 0.22 µm membrane filter and pH was adjusted to 7.2-7.4. Prior to use, it was kept in 5% CO2 incubator at 39° C for at least 2 hours for equilibration. Slicing method was used for sufficient recovery of oocytes. Ovaries were placed one by one in an oocyte-searching dish containing oocyte collection medium. Then the ovaries were grasped by sterilized forceps and the surface visible all follicles (3-6 mm diameter) were punctured by slicing with sterile scalpel. At a time 10-12 ovaries were processed in this way and the dish was observed under zoom stereomicroscope under 20 X magnification. For oocyte pick up sterilized Pasteur pipettes were used. They were pulled over the flame to make the inner diameter of 300 µm. Oocytes along with the cumulus cells (COCs) were picked up gently without damaging the cumulus cells and were kept in a 35 mm Petri dish containing washing medium for further washing.

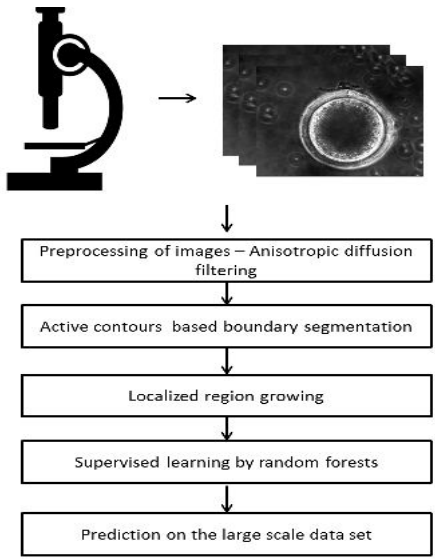

Figure 1. Proposed Workflow and the machine learning pipeline

In the current protocol adopted, the collected oocytes are graded A, B, C and D according to the number of cumulus cell layers surrounding the oocytes and the homogeneity of ooplasm as shown in Figure 2. Textural features of the image are difficult to analyze via visual inspection. Since there are a lot of image filters which can be applied to obtain feature banks, a smart selection of features is important for fast classification with uncompromised accuracy.

### B. Problem Statement

Given a set of images $I\_train$, a set of features can be extracted at a particular position $x$. A feature matrix is generated with corresponding image class is stacked adjacent to the features. Hence each row represents a sample for training the classification algorithm. The prediction for each image can be represented as $P(c|I(x); I_{train})$, where $c = \{A, B, C, D\}$. The final assigned class to an image can be represented as:

$$c_{assigned} = \max\{P(c|I(x); I_{train})\} = \max\{P_{trained}\}$$

where, $P_{trained}$ represents set of all $P$'s for all the pixels of an image. Hence the class with maximum votes is assigned to the image. The classification of an image falls into one of the 4 classes and hence this is a multi-class classification problem.

The samples fall through an ensemble of such decision trees and this paradigm is the well-known Random Forest (RF).

### C. Random Forest based classification

For an introduction to RF, the reader is directed to the detailed treatment in [8]. The strength of random forest lies in its robustness to noise and ability to calculate and tune feature importance. In a similar classification task in [9], the authors reported the accuracy of random forest with respect to support vector machines and back propagation neural networks. Random forest outperforms these classification algorithms in cases of unbalanced, multi-class and small sample data. Hence RF is suitable for our classification problem. The output of each tree in a classification problem is the probability that a particular label is assigned to an incoming observation. The prediction of the random forest is the average of all such decision trees. Taking the average of the predictions among all the trees reduces the variance of the prediction and reduces the bias of one particular decision tree [10]. Random Forests are very robust and can be computationally efficient in higher dimensions and can be parallelized [11]. The decision making workflow of the random forest algorithm is summarized below:

- For each class $c \in C$, the posterior probability of class $c$ given observation at $x$, is calculated. The probability is given by $p_t(c|I(x))$ for $t = 1, \ldots, T$, $T$ being the number of trees using for training.
- The weighted average $p_{avg}$ over selected trees is calculated as: $p_{avg}(c|I(x)) = \frac{\sum_{n=1}^{T}(a_t)p_t(c|I(x))}{\sum_{n=1}^{T}(a_t)}$ (2)

$$\forall a > 0$$

Parameters such as depth of trees and number of leaf nodes can be tuned to obtain optimal performance. The discussion of the influence of these parameters is out of the scope of this paper. For the extraction of relevant features, an object detection segmentation algorithm has to be applied.

*D. Boundary detection and nucleus segmentation*

The contents of the image which form the basis of the classification problem are the nucleus and the adjacent COC layer. The different classes of the oocytes depend on the content in the COCs. The difference in width of the nucleus and the entire cell is a varying factor in all groups and we decided to segment the nucleus and identify the boundary of the cell. Hence two concentric regions have to be identified. One is the boundary delineation from the background and the second being the nucleus region. We fit parametric curves for both these regions to give us useful measures such as relative area occupied by the cell. The images are pre-processed with an anisotropic diffusion based Gaussian filter which preserves edges. It has been shown that anisotropic diffusion based Gaussian filtering can act as an edge preserving filter [12]. Additionally, image correction methods (e.g. auto white balancing) can be employed to ensure exact color representations [13]. This is essential for the performance of the snakes. For the outer boundary segmentation, we use the classical active contours provided by Kass et al. [14].

Following the outer boundary detection, we have to detect the nucleus of the cell which was found to have a distinct intensity from the rest of the cell area. Hence we applied a localized region growing (RG) algorithm (constraint based region growing with shape fitting) [15]. The seed point for the region growing algorithm was selected automatically as the center of the boundary enclosing circle from previous stage. This turned out to be a very good initial seed point guess. The threshold value for RG was selected heuristically.

*C. Feature Selection*

In general, we can represent the microscopic images with two kinds of features: Texture based and Contour based [10]. Texture based features which were used include traditional image features such as gradient of the image, laplacian of Gaussian, Haar and LBP. The contour based features are inferred from the segmentation process. Analyzing the different features provided by segmentation process, most important were: (i) area occupied by the two circles, (ii) distance from the image boundary, (iii) intensities in the cumulus layer, and (iv) strength of edge linking. For example, in grades A and D the edges occur in small connected components whereas in grades B and C the structures form a continuous cumulus layer. This agrees with the fact that grades A and D have smaller circular structures in the COC. Even though other features can be extracted based on different filters applied on the images, the performance of the machine learning algorithm does not always improve with increase in the number of features and this is often referred to as the "Curse of dimensionality" [16].

## III. RESULTS AND DISCUSSION

A total of 80 oocytes with almost equal distribution of classes were obtained for this work. Segmentation task was carried out for the ground truth data which comprised of around 30 datasets. In Fig. 2 we illustrate the different grades of the oocyte and its corresponding segmented boundaries obtained using active contours (cellular) and region growing (nucleus). Fig. 3 shows the performances of the segmentation algorithm vis-à-vis ground truth obtained by manual segmentation. The dice coefficient and the Rand index of the two concentric circles are shown; more details of the measures are available in [16].

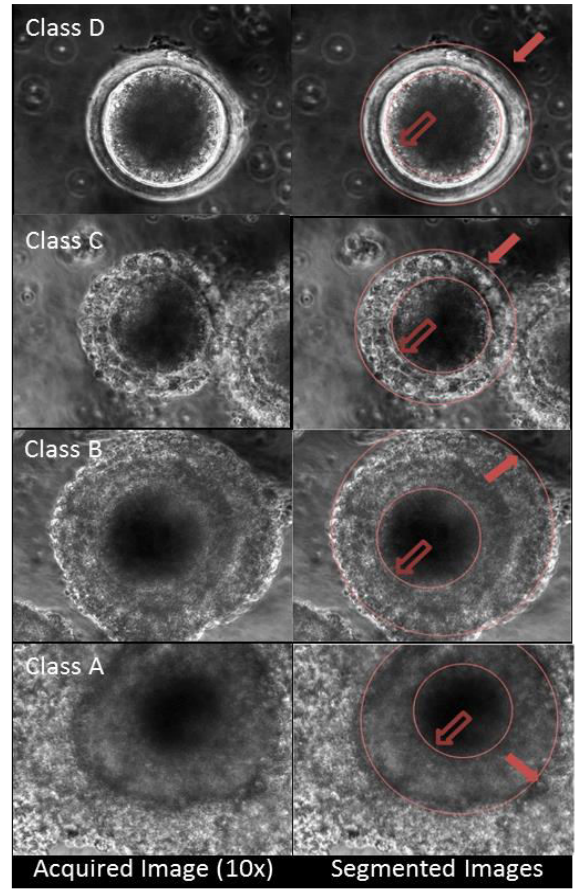

Figure 2. Parametric curve fit based on region growing for nuclear segmentation (inner circle, *marked with hollow red line*)and active contours for cellular boundary segmentation (outer circle, *marked with solid red line*) shown for various classes of Cumulus Oocyte Complexes (COCs)

*A. Random Forest classification*

The Matlab (Mathworks Inc.) implementation of Random Forest, called the Treebagger was made use of in the classification problem. As mentioned, texture based and contour based features were extracted amounting to a total of 31 features. Fig. 4a shows a plot of the importance of the features based on the out of bag errors for the classification. The first 5 set of features which appear to be more important than the rest are the ones which were extracted based on the segmentation. To name a few, the features were the radius of the inner and outer circles, and the ratios and the area occupied by the entire cell in the fixed field of view. For the purpose of training the random forest, we started with a 30-70 approach at the beginning, that is, we maintained 30% of the datasets for training and the rest for testing. On further analysis we found that increasing the training set did not improve the accuracy to a considerably. The testing images had approximately equal distribution of the classes of oocytes.

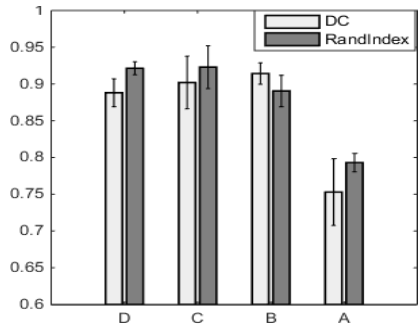

Figure 3. Performances of image segmentation for each class of Oocytes

The final accuracy obtained across different number of trees as well as different depth of trees was found to be 88.35% with all the images belonging to grade D and grade A being classified correctly, and misclassification remained only between grade B and grade C images. If only texture based features are used, the RF resulted in an accuracy of 75%, showing the importance of contour based features. Hence in future more contour based features can be investigated. The improvement in segmentation can further improve the quality of features used for random forest classification, and in turn improve the overall prediction accuracy.

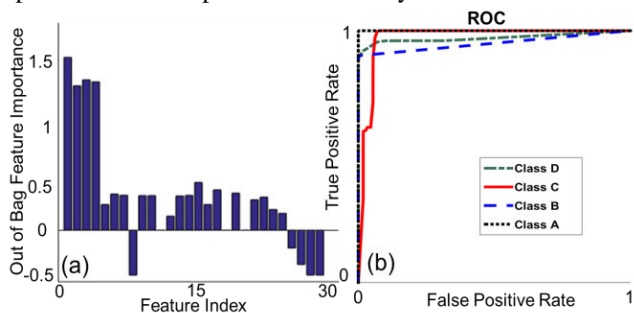

Figure 4. Efficacy of selected features (a) and it's the efficiency of classification of oocytes into four constituent classes (b).

## IV. CONCLUSION

The article illustrates a semi-automatic framework for predicting the class an oocyte, which has so far been done by visual inspection by biologists/embryologists. The framework is based on active contour and region-growing segmentation on the acquired microscopic images. The framework discussed is static in nature in that it works on a set of images that are fed into the framework. However, an extension to dynamic evaluation of the oocytes can be made and that would only increase the accuracy of the framework. The machine learning subsystem consists of feature selection and utilized a random forest decision method for classification yielding an accuracy of 88%. Interestingly, very little attention has been invested so far, to improve fertilization performance solely by optimization of oocyte selection performance using image analysis and machine learning. Although other classifiers like SVM and Neural Networks could have been used, RF was chosen due to applicability to a large number of datasets and parallelization capability. These methods can potentially improve IVF performance in mammalian species, and currently being used effectively for porcine oocyte selection by livestock biotechnology researchers.